\pdfoutput=1

\documentclass[11pt]{article}

\usepackage{acl}

\usepackage{times}
\usepackage{latexsym}

\usepackage{booktabs}
\usepackage{caption}
\usepackage{subcaption}
\usepackage[pdftex]{graphicx}
\usepackage{url}

\usepackage{xstring}
\usepackage{color, colortbl}
\usepackage{xcolor}
\usepackage{collcell}
\usepackage{multirow}
\usepackage{pgf}
\usepackage{pdflscape}
\usepackage{adjustbox}
\usepackage{float}
\usepackage{diagbox}
\usepackage{enumitem}

\definecolor{darkred}{rgb}{0.0, 0.3, 0.6}

\newcommand{\ApplyRedGradient}[1]{%
    \IfBeginWith{#1}{0.}{%
        \pgfmathsetmacro{\PercentColor}{70.0*#1}%
            \edef\x{\noexpand\cellcolor{darkred!\PercentColor}}\x\textcolor{black}{#1}%
    }%
    {%
        \IfBeginWith{#1}{1.}{%
            \pgfmathsetmacro{\PercentColor}{65.0*#1}%
                \edef\x{\noexpand\cellcolor{darkred!\PercentColor}}\x\textcolor{black}{#1}%
        }%
        {%
          \pgfmathsetmacro{\PercentColor}{0.001}%
          #1%
        }%
    }%
}
\newcommand{\ApplyGradientScaled}[1]{%
    \IfBeginWith{#1}{0.}{%
        \pgfmathsetmacro{\PercentColor}{(40.0*#1)}%
            \edef\x{\noexpand\cellcolor{darkred!\PercentColor}}\x\textcolor{black}{#1}%
    }%
    {%
      \pgfmathsetmacro{\PercentColor}{0.001}%
      #1%
    }%
}

\newcolumntype{R}{>{\collectcell\ApplyRedGradient}r<{\endcollectcell}}

\usepackage[T1]{fontenc}

\usepackage[utf8]{inputenc}

\usepackage{microtype}

\usepackage{twemojis}

\title{Prompting open-source and commercial language models for \\grammatical error correction of English learner text}

\author{
    {\bf Christopher Davis} \texttwemoji{grapes} ~~~~
    {\bf Andrew Caines} \texttwemoji{grapes} ~~~~
    {\bf {\O}istein Andersen} \texttwemoji{grapes} ~~~~ \\
    {\bf Shiva Taslimipoor} \texttwemoji{grapes} ~~~~ 
    {\bf Helen Yannakoudakis} \texttwemoji{hibiscus} ~~~~ 
    {\bf Zheng Yuan} \texttwemoji{hibiscus} ~~~~ \\
    {\bf Christopher Bryant} \texttwemoji{sunflower} ~~~~
    {\bf Marek Rei} \texttwemoji{palm_tree} ~~~~
    {\bf Paula Buttery} \texttwemoji{grapes} ~~~~ \\
    \texttwemoji{grapes} ALTA Institute, Computer Laboratory, University of Cambridge, U.K. \\
    \texttwemoji{sunflower} Writer Inc., San Francisco, California, U.S.A. \\
    \texttwemoji{hibiscus} King's College London, U.K. \\
    \texttwemoji{palm_tree} Imperial College London, U.K. \\
    \small \texttwemoji{grapes} \texttt{firstname.secondname@cl.cam.ac.uk}
    \hspace{4mm}
    \small \texttwemoji{sunflower} \texttt{cjb255@cam.ac.uk} \\
    \small \texttwemoji{hibiscus} \texttt{firstname.secondname@kcl.ac.uk}
    \hspace{4mm}
    \small \texttwemoji{palm_tree} \texttt{marek.rei@imperial.ac.uk}}

\begin{document}
\maketitle
\begin{abstract}
Thanks to recent advances in generative AI, we are able to prompt large language models (LLMs) to produce texts which are fluent and grammatical. In addition, it has been shown that we can elicit attempts at grammatical error correction (GEC) from LLMs when prompted with ungrammatical input sentences. We evaluate how well LLMs can perform at GEC by measuring their performance on established benchmark datasets. 
We go beyond previous studies, which only examined GPT$*$ models on a selection of English GEC datasets, by evaluating seven open-source and three commercial LLMs on four established GEC benchmarks. 
We investigate model performance and report results against individual error types. Our results indicate that LLMs do not always outperform supervised English GEC models except in specific contexts -- namely commercial LLMs on benchmarks annotated with fluency corrections as opposed to minimal edits. We find that several open-source models outperform commercial ones on minimal edit benchmarks, and that in some settings zero-shot prompting is just as competitive as few-shot prompting.
\end{abstract}

\section{Introduction}

Grammatical error correction (GEC) of second language learner English text is an important task in Educational AI.
Its main applications include: i) enabling learners to receive instant feedback on their written work, ii) providing features for automarking, and iii) profiling learners' grammatical knowledge in such a way as to facilitate personalised learning \cite{andersen-etal-2013-developing,yannakoudakis2018developing,zaidiaccurate}.

There is a long history of GEC research in the field of computational linguistics, developing from rule-based to statistical approaches to neural network models, as has happened with other tasks in natural language processing \cite{bryant-et-al-2023}. Given the recent emergence of large language models (LLMs), such as Open\-AI's GPT$*$ and Meta's Llama LLMs, 
it is natural to ask how well they can perform at GEC and how they compare to existing state-of-the-art supervised approaches \cite{caines2023application}.

To answer this question, we aim to elicit minimal edit style corrections from LLMs through zero-shot and few-shot prompting.
Minimal edit correction of text aims to resolve any grammatical errors in a text while staying as close as possible to the phrasing and lexical choices of the original. This is sometimes held up as a distinct alternative to fluency correction -- where texts are rewritten for naturalness -- though in reality the two annotation methods are not completely separable from each other \cite{bryant-et-al-2023}.
The tendency so far has been to annotate GEC datasets with minimal edits only, and so that is the way past systems have been trained. This is an important point to note, as LLMs by default will output a transformative fluency correction of ungrammatical text \cite{coyne2023analyzing,fang2023chatgpt,loem-etal-2023-exploring}. 
In this paper, we attempt to prompt LLMs to perform minimal edit correction rather than fluency correction, so that the outputs are comparable to previous systems.
Minimal edit corrections are additionally valuable within an educational setting, for example displaying grammatical errors to a learner \cite{yannakoudakis2018developing}.

We evaluate three commercial and seven open source LLMs on four publicly available GEC benchmarks: CoNLL 2014 \cite{ng-etal-2014-conll}, the FCE Corpus \cite{yannakoudakis-etal-2011-new}, JFLEG \cite{napoles-etal-2017-jfleg}, and Write\&Improve + LOCNESS (W\&I) from the BEA-2019 shared task \cite{bryant-etal-2019-bea}.
This builds on previous work, which involved GPT$*$ models only and at most three GEC benchmarks \cite{coyne2023analyzing,fang2023chatgpt,loem-etal-2023-exploring}.
We find that some of our chosen open-source models outperform GPT-3.5 Turbo on English GEC benchmarks which have been annotated in a minimal edit fashion. In contrast, GPT-3.5 performs best on a test set annotated with fluency corrections. 

We evaluate several zero-shot and few-shot prompts, but find that different models require different styles of prompting. Some models appear to respond better to few-shot prompting than others, and certain prompt templates work best with a specific LLM and dataset rather than universally across the board. We provide empirical evidence that prompting LLMs does not outperform existing state-of-the-art, supervised GEC models -- though the search space over LLMs, prompt templates, and few-shot learning is so great that our results can only be considered a building block in the full picture of LLM evaluation on English GEC. 
Our investigations can serve as a comparison point for future work on GEC with LLMs.
We make our code, prompts, few-shot examples and model predictions publicly available.\footnote{\url{https://github.com/chrisdavis90/gec-prompting-public}}



\section{Related work}

\paragraph{Grammatical error correction (GEC).} GEC is the task of returning an edited version of an input text such that any grammatical errors are corrected. It is a longstanding task in research on NLP for educational applications. 


Thanks to large-scale annotation projects and the public release of labelled data, GEC systems can be built aiming at \emph{general} correction of all error types. Machine translation GEC systems pioneered this general purpose approach, at first with statistical and later neural models \cite{brockett-etal-2006-correcting,junczys-dowmunt-etal-2018-approaching,yuan-bryant-2021-document}. More recently, edit-based approaches have been proposed in which corrections are applied on a sequence labelling \cite{omelianchuk-etal-2020-gector} or sequence-to-sequence basis \cite{stahlberg-kumar-2020-seq2edits}. \newcite{bryant-et-al-2023} offers a comprehensive survey of the history and current state of GEC.

\paragraph{GEC with large language models (LLMs).}
The term `large language model' is currently used to refer to a variety of neural networks developed by a number of organisations and businesses. These models reached the mainstream media through GPT-3 and ChatGPT, and as a result there is now a widespread awareness of `generative AI' -- in particular relating to text generation -- amongst the general public. Open\-AI's GPT$*$ models feature in this paper, alongside others for comparison. We have selected ten open-source and proprietary LLMs for reasons described in \autoref{sec:models}.

Recent studies indicate that LLMs from Open\-AI can be prompted to generate corrected versions of ungrammatical inputs. \newcite{wu2023chatgpt} compares ChatGPT to Grammarly and GECToR \cite{omelianchuk-etal-2020-gector} on a sample of 100 sentences from the CoNLL-14 test set. \newcite{coyne2023analyzing} compares GPT-3.5 and GPT-4\footnote{Specifically the \texttt{text-davinci-003} GPT-3.5 model, and a GPT-4 model \texttt{gpt-4-0314}.} to two GEC systems on English benchmarks \cite{yasunaga2021lm, liu2021neural}, whilst \citet{fang2023chatgpt} compare ChatGPT with multiple baselines including TagGEC \cite{stahlberg-kumar-2021-synthetic} and T5 \cite{rothe-etal-2021-simple}. Finally \citet{loem-etal-2023-exploring} compare ChatGPT and GPT-3.5 (\texttt{text-davinci-003}) with models trained on synthetic data \cite{grundkiewicz-junczys-dowmunt-2019-minimally,grundkiewicz-etal-2019-neural}. 

\citet{coyne2023analyzing,fang2023chatgpt,loem-etal-2023-exploring} all perform evaluation of English GEC on some combination of the JFLEG, CoNLL-14 and W\&I+LOCNESS test sets. 
They find that the GPT$*$ models set new state-of-the-art (SOTA) performance on the JFLEG dataset in which the annotators were permitted to carry out naturalistic fluency rewrites \cite{napoles-etal-2017-jfleg}.
However, they perform worse than SOTA on CoNLL-14 and W\&I+LOCNESS, which are much larger, more popular datasets that were annotated on the basis of minimal edit corrections.
\citet{coyne2023analyzing} and \citet{fang2023chatgpt} found through further investigation that the GPT$*$ models have a tendency to over-correct and make extraneous fluency edits. This explains why it is that they can score so highly on JFLEG but not on minimal edit data.
\citet{loem-etal-2023-exploring} meanwhile investigated the possibility of prompting for minimal edits rather than fluency rewrites, and obtained promising improvements which motivate further work.


Both \newcite{wu2023chatgpt} and \newcite{coyne2023analyzing} additionally carried out human evaluation to rate the output from each system and found a preference amongst human raters for the GPT$*$ outputs because they were considered to be more fluent. They also found instances of \emph{under}-correction in the reference sentences derived from human annotators: in other words the LLMs were able to catch and correct errors which had not been corrected by the original annotators. These human evaluations are tentative only, since they involve only small samples of 100 sentences at a time from each test set. 

While more fluent corrections may be preferred by human evaluators, they may not aid language learners if they diverge too greatly from the original text. Existing annotation guidelines for error correction state that edits should be as minimal as possible so that the learner can be helped to express what they are trying to say, rather than told how to express it differently (which may otherwise discourage them); i.e.~how to amend an error rather than avoid it \cite{nicholls-2003}.
Consequently, although both minimal and fluent corrections may be valuable to different user groups, we focus on minimal corrections for educational applications in this paper.

\begin{table}
  \small
  \centering
  \begin{tabular}{llrr}
    \toprule
    Dataset     & Split & \# Tokens & \# Sentences \\
    \midrule
    CoNLL-14 & Test & 30,144 & 1,312 \\
    W\&I+LOCNESS & Dev & 86,973 & 4,384 \\
    FCE & Dev & 34,748 & 2,191 \\ 
    & Test & 41,932 & 2,695 \\
    JFLEG & Dev & 14,010 & 754 \\
    & Test & 14,096 & 747 \\
    \midrule
    W\&I+LOCNESS & Train & 628,719 & 34,308 \\
     & Sampled & 18,386 & 1,000\\
    FCE & Train & 454,736 & 28,350 \\
     & Sampled & 16,112 & 1,000 \\
    \bottomrule
  \end{tabular}
  \caption{\label{table:gec_datasets} Grammatical error correction datasets. }
\end{table}

\section{Datasets}

We compare model performance on four publicly available and well-known English language GEC datasets: CoNLL-14 Test \cite{ng-etal-2014-conll}, JFLEG Dev and Test \cite{napoles-etal-2017-jfleg}, FCE Dev and Test \cite{yannakoudakis-etal-2011-new}, and W\&I+LOCNESS Dev\footnote{The test set for W\&I+LOCNESS is not publicly available.} \cite{bryant-etal-2019-bea}. We additionally sample 2,000 sentences uniformly from FCE train and W\&I+LOCNESS train to construct a development set in order to filter the set of prompt templates. \autoref{table:gec_datasets} presents the number of sentences and tokens per dataset.

The CoNLL-14 test set contains 50 essays written by undergraduate students at the National University of Singapore on one of two topics. It has featured in multiple GEC studies, and new SOTA performance was reported by \citet{zhou-et-al-2023-improving-gec} in a recent paper describing decoding interventions.

JFLEG dev and test contain approximately 1.5k sentences randomly sampled from essays by learners of English of unknown proficiency levels, and corrected by crowdworkers. Annotators were permitted to make fluency corrections to the sentences: not just minimal edits for grammaticality.
\citet{stahlberg-kumar-2021-synthetic} achieved the current SOTA performance on JFLEG for a single system with their guided approach to synthetic generation of training data based on error type distributions found in annotated corpora.

FCE dev and test feature essays written by intermediate learners of English (CEFR levels B1 and B2). It is a subset of the Cambridge Learner Corpus and has also been used in multiple GEC studies. Current SOTA was established by \citet{yuan-bryant-2021-document} with a multi-encoder model which encodes a given sentence and the preceding one separately, integrating them in the decoder.

W\&I+LOCNESS is a hybrid dataset made up of native speaker essays written by undergraduate students (LOCNESS; \citet{locness}) and essays submitted to the Write\&Improve learning platform by learners of English at varying levels of proficiency (W\&I). It was prepared for the BEA 2019 Shared Task on GEC \cite{bryant-etal-2019-bea}, and SOTA was achieved by \citet{qorib-etal-2022-frustratingly} with system combination across multiple GEC models.

Each dataset was processed with ERRANT \cite{felice-etal-2016-automatic,bryant-etal-2017-automatic}, an automatic error annotation tool, in order to be standardised into a common format. Consequently the datasets are in tokenised M2 format, and we first need to detokenise them as LLMs expect untokenised inputs. To carry out this task, we use the Moses detokeniser\footnote{\url{https://github.com/luismsgomes/mosestokenizer}} and a rule-based heuristic to combine negative contractions which are not fully handled by the detokeniser\footnote{The detokeniser transforms token sequences such as ``couldn 't'' to ``couldn't'' in a satisfactory manner but sequences such as ``could n't'' are missed.}.

\begin{table*}[t]
    \footnotesize
    \centering
    \begin{tabular}{lp{0.8\linewidth}}
        \toprule
        Name & Prompt \\
        \midrule
        \textsc{min} & Make minimal changes to the following text such that it is grammatically correct. \{text\}\\
        \textsc{elt}$^{\dagger}$ & You are an English language teacher. A student has sent you the following text. \textbackslash n\{text\}\textbackslash nProvide a grammatical correction for the text, making only necessary changes. Do not provide any additional comments or explanations. If the input text is already correct, return it unchanged. \\
        \textsc{tool}$^{*\dagger}$ & You are a grammatical error correction tool. Your task is to correct the grammaticality and spelling in the input sentence. Make the smallest possible change in order to make the sentence grammatically correct. Change as few words as possible. Do not rephrase parts of the sentence that are already grammatical. Do not change the meaning of the sentence by adding or removing information. If the sentence is already grammatically correct, you should output the original sentence without changing anything. \textbackslash n\textbackslash nInput sentence: \{text\}\textbackslash nOutput sentence: \\
        
        \textsc{dn} & Please correct the following text. Do not attempt to rewrite it into perfect English or to interpret the text. Often, things could be expressed better by paraphrase, but the task is to make minimal changes to correct the text.  Do not change anything that is correct.  Please make no changes if there are no errors.  \\
        \textsc{cyn}$^{\dagger}$ & Reply with a corrected version of the input sentence with all grammatical and spelling errors fixed. If there are no errors, reply with a copy of the original sentence.\textbackslash n\textbackslash nInput sentence: \{text\}\textbackslash nCorrected sentence:  \\
        \textsc{con} & This sentence is ungrammatical: \{text\}. I would correct the sentence with as few changes as possible like this:  \\
        \bottomrule
    \end{tabular}
    \caption{\label{table:prompts_brief} The set of prompts used in zero- and few-shot settings.
    $^{*}$There are two versions of the \textsc{tool} prompt: with and without quotations around the \{text\}. $^{\dagger}$ indicates prompts used in few-shot evaluation.}
\end{table*}

\section{Models}
\label{sec:models}

We evaluate three commercial and seven open-source LLMs. We include more open-source than commercial models as we assume that the latter will have a performance advantage and wish to investigate whether open-source models can perform in comparable ways. If so, this would be positive news from an open-science perspective.
For the commercial LLMs, we include Open\-AI's GPT-3.5-turbo and GPT-4 models \cite{OpenAI2023gpt}, and Cohere's Command model.\footnote{Both GPT-3.5-turbo and GPT-4 are the 0613 versions.
Cohere's Command is ``v1''.}
Many more are available but due to budget constraints we work only with these three and we only evaluate GPT-4 in the zero-shot setting. We choose not to work with ChatGPT as it has been engineered to function as a chatbot. 

For the open-source models, we select instruction-tuned models because the majority of our prompt templates contain instructions, and we evaluate the largest model from each model type that fits on a server with two A100 80GB NVIDIA GPUs. Our upper bound on model size relates to the computing resources available to us at the time of writing.

The open-source models are: OPT-IML-Max-30B \cite{iyer2022opt}, Llama-2-70B-chat \cite{touvron2023llama}, Stable Beluga 2 \cite{StableBelugaModels}, Falcon-40B-Instruct \cite{falcon40b}, Flan-T5-XXL \cite{chung2022scaling}, BLOOMZ-7B1 \cite{muennighoff2022crosslingual}, InstructPalmyra-20B \cite{InstructPalmyra}. This is a representative sample of the models available, with a range of sizes and architectures. Approximate model sizes are given in \autoref{table:models} in the Appendix.
We use HuggingFace \cite{wolf-etal-2020-transformers} to run the models and load them with float16 precision.\footnote{We use bfloat16 for Falcon-40B-instruct.} 

\subsection{Prompting LLMs for grammatical error correction}
\label{section:prompt_templates}

    

Prior work has shown that prompt format and wording can have a significant impact on task performance \cite{jiang-etal-2020-know,shin-etal-2020-autoprompt,schick-schutze-2021-just}. We therefore evaluate and compare models across a selection of prompt templates (hereinafter prompts). In order to constrain the scope of experiments, we carry out two filtering and evaluation steps to construct and evaluate a set of zero- and few-shot prompts as follows.



We collect eleven zero-shot prompts based on a survey of NLP colleagues and related work.\footnote{We considered a wide set of prompts used in related work but ultimately decided against their inclusion due to the estimated difficulty in replication and time/budget constraints.} 
We first evaluate the zero-shot prompts with each model on a development set of 2,000 sentences sampled uniformly from the FCE and W\&I+LOCNESS training sets. From these results we exclude four prompts with the lowest maximum scores, leaving seven prompts to evaluate in the zero-shot setting on the three development datasets: FCE, JFLEG, and W\&I+LOCNESS.\footnote{Details are provided in \autoref{appendix:filtering_zero_shot_prompts}.} 

We then select the 3 best-performing zero-shot prompts and create few-shot versions using 1, 2, 3, and 4 examples -- 12 few-shot prompts in total. While related work samples few-shot examples, we make the decision to use a fixed set and order of examples to control the experimental parameters. A dynamic set of few-shot examples would require multiple samples per sentence in order to obtain a clear view of few-shot performance for each model. In addition, we evaluate models using the best performing few-shot prompt from \citet{coyne2023analyzing}. 

\autoref{table:prompts_brief} lists the prompts we evaluate in zero- and few-shot settings (see \autoref{app:prompt_templates} for the complete set). Briefly, prompts \textsc{min} and \textsc{dn} contain general instructions to make minimal corrections, prompt \textsc{elt} uses an ``English language teacher'' expert, \textsc{tool} uses a ``grammatical error correction tool'' expert, \textsc{cyn} is the prompt from \citet{coyne2023analyzing}, and \textsc{con} frames the GEC instruction as a continuation. The set of few-shot examples are listed in Appendix \autoref{table:few_shot_examples}.

\begin{table*}[t]
    \centering
    \small
    \begin{tabular}{l|r|r|l|r|r|l|r|r|l}
        \toprule
         & \multicolumn{3}{c|}{FCE$_{dev}$} & \multicolumn{3}{c|}{JFLEG$_{dev}$} & \multicolumn{3}{c}{W\&I$_{dev}$} \\
        Model & F$_{0.5}$ & N & Prompt & GLEU & N & Prompt & F$_{0.5}$ & N & Prompt \\
        \midrule
BLOOMZ & 0.349 & 3 & \textsc{cyn} & 0.456 & 2 & \textsc{cyn}$^{\dagger}$ & 0.347 & 3 & \textsc{cyn} \\
FLAN-T5 & 0.447 & 1 & \textsc{tool} & 0.463 & 1 & \textsc{tool} & 0.423 & 3 & \textsc{tool} \\
InstructPalmyra & 0.341 & 2 & \textsc{cyn} & 0.517 & 0 & \textsc{tool} & 0.374 & 2 & \textsc{cyn} \\
OPT-IML & 0.395 & 0 & \textsc{tool} & 0.506 & 2 & \textsc{cyn}$^{\dagger}$ & 0.400 & 3 & \textsc{elt} \\
Falcon-40B-Instruct & 0.425 & 2 & \textsc{tool} & 0.548 & 4 & \textsc{cyn} & 0.454 & 4 & \textsc{tool} \\
Llama 2 & 0.323 & 0 & \textsc{tool} & 0.500 & 0 & \textsc{tool} & 0.359 & 0 & \textsc{tool} \\
Stable Beluga 2 & 0.403 & 0 & \textsc{tool} & 0.563 & 0 & \textsc{cyn} & 0.447 & 0 & \textsc{tool} \\
Command & 0.353 & 0 & \textsc{tool} & 0.543 & 2 & \textsc{cyn}$^{\dagger}$ & 0.391 & 0 & \textsc{tool} \\
GPT-3.5 Turbo 0613 & 0.416 & 0 & \textsc{elt} & 0.577 & 4 & \textsc{tool} & 0.439 & 1 & \textsc{tool} \\
GPT-4 0613$^{*}$ & \textbf{0.474} & 0 & \textsc{elt} & \textbf{0.582} & 0 & \textsc{tool} & \textbf{0.510} & 0 & \textsc{tool} \\
\midrule

C: GPT 3.5 text-davinci-003 & \multicolumn{1}{c|}{--} & \multicolumn{1}{c|}{--} & \multicolumn{1}{c|}{--} & 0.582 & 0 & \multicolumn{1}{c|}{--} & \multicolumn{1}{c|}{--} & \multicolumn{1}{c|}{--} & \multicolumn{1}{c}{--} \\

C: GPT 3.5 text-davinci-003 & \multicolumn{1}{c|}{--} & \multicolumn{1}{c|}{--} & \multicolumn{1}{c|}{--} & 0.590 & 2 & \multicolumn{1}{c|}{--} & \multicolumn{1}{c|}{--} & \multicolumn{1}{c|}{--} & \multicolumn{1}{c}{--} \\

C: GPT-4 0314 & \multicolumn{1}{c|}{--} & \multicolumn{1}{c|}{--} & \multicolumn{1}{c|}{--} & \bf 0.601 & 0 & \multicolumn{1}{c|}{--} & \multicolumn{1}{c|}{--} & \multicolumn{1}{c|}{--} & \multicolumn{1}{c}{--}  \\

C: GPT-4 0314 & \multicolumn{1}{c|}{--} & \multicolumn{1}{c|}{--} & \multicolumn{1}{c|}{--} & 0.600 & 2 & \multicolumn{1}{c|}{--} & \multicolumn{1}{c|}{--} & \multicolumn{1}{c|}{--} & \multicolumn{1}{c}{--}  \\

\bottomrule
    \end{tabular}
\caption{\label{table:final_zero_few_shot_max_dev} Results on the FCE, JFLEG, and W\&I dev sets, using the best prompt per model. ``N'' refers to the number of few-shot examples. ``Prompt'' refers to the type of prompt instruction: \textsc{tool} is the GEC tool expert, \textsc{elt} the English Language Teacher expert, \textsc{cyn} refers to the prompt from \citet{coyne2023analyzing} with our few-shot examples, and \textsc{cyn}$^{\dagger}$ indicates the template with their few-shot examples. Performance reported in previous work is shown in the lower part of the table. C: refers to \citet{coyne2023analyzing}. GPT-4$^{*}$ was only evaluated in a zero-shot setting.}
\end{table*}

\paragraph{Generation hyper-parameters} We use the following settings for all models -- we set temperature to 0.1, top-K to 50, and top-P to 1.0. Preliminary work has shown that lower temperature values result in better GEC performance \cite{coyne2023analyzing}, and importantly, we want the model to make minimal edits and stay as close as possible to the original sentence. For some models the lowest temperature is 0.1, and so we set the parameter to this value to be constant across all models.

\begin{table}[h]
\footnotesize
\centering
\begin{tabular}{lRRR}
\toprule
Model & \multicolumn{1}{l}{Prec} & \multicolumn{1}{l}{Rec} & \multicolumn{1}{l}{F$_{0.5}$} \\
\midrule    
BLOOMZ & 0.475 & 0.169 & 0.349 \\
FLAN-T5 & 0.615 & 0.213 & 0.447 \\
InstructPalmyra & 0.357 & 0.287 & 0.341 \\
OPT-IML & 0.559 & 0.182 & 0.395 \\
Falcon-40B-Instruct & 0.438 & 0.381 & 0.425 \\
Llama 2 & 0.304 & 0.428 & 0.323 \\
Stable Beluga 2 & 0.396 & 0.432 & 0.403 \\
Command & 0.356 & 0.342 & 0.353 \\
GPT-3.5 Turbo 0613 & 0.398 & 0.504 & 0.416 \\
GPT-4 0613 & 0.473 & 0.477 & 0.474 \\

\bottomrule
\end{tabular}
\caption{\label{table:precision_recall_fce_dev} Performance for models on the FCE development set, using their best prompts -- models ordered by increasing size. Prec/Rec are precision and recall, respectively.}
\end{table}

\paragraph{Evaluation}

As per the recommendations in \citet{bryant-et-al-2023}, we evaluate the FCE and W\&I corpus in terms of F$_{0.5}$ using ERRANT \cite{bryant-etal-2017-automatic}, the CoNLL-2014 test set in terms of F$_{0.5}$ using the M$^2$ scorer \cite{dahlmeier-ng-2012-better}, and the JFLEG corpus using GLEU \cite{napoles-etal-2015-ground}.

Along with the open search space in prompt design, a practical question arises as to how much time and effort to dedicate to implement a model- or prompt-specific post-processing step to extract the generated hypothesis sentence from the model output. Due to the number and variety of models and prompts, it's possible that each model--prompt combination will generate a different output format, and clearly, the quality of the post-processing step will impact evaluation measures. For all models, we replace all new line tokens with blank spaces, replace sequences of multiple spaces with a single space, and remove all trailing quotation marks. We also remove strings from the start and end of sentences based on keyword matching -- for example, we remove ``Output sentence: '', ``Corrected sentence: '', and ``Input sentence: '' from the start of sentences. The output from Llama-2-chat was particularly noisy and required more rules -- we detail our processing steps in \autoref{appendix:post_processing_details}.

\begin{table*}[t]
    \centering
    \small
    \begin{tabular}{l|r|r|l|r|r|l|r|r|l}
        \toprule
         & \multicolumn{3}{c|}{FCE$_{test}$} & \multicolumn{3}{c|}{JFLEG$_{test}$} & \multicolumn{3}{c}{CoNLL-14$_{test}$} \\
        Model & F$_{0.5}$ & N & Prompt & GLEU & N & Prompt & F$_{0.5}$ & N & Prompt \\
        \midrule

BLOOMZ & 0.358 & 3 & \textsc{cyn} & 0.498 & 2 & \textsc{cyn}$^{\dagger}$ & 0.405 & 3 & \textsc{cyn} \\
FLAN-T5 & \textbf{0.463} & 1 & \textsc{tool} & 0.508 & 1 & \textsc{tool} & 0.397 & 3 & \textsc{tool} \\
InstructPalmyra & 0.396 & 2 & \textsc{cyn} & 0.572 & 0 & \textsc{tool} & 0.499 & 2 & \textsc{cyn} \\
OPT-IML & 0.400 & 0 & \textsc{tool} & 0.521 & 2 & \textsc{cyn}$^{\dagger}$ & 0.396 & 3 & \textsc{elt} \\
Falcon-40b-Instruct & 0.456 & 2 & \textsc{tool} & 0.602 & 4 & \textsc{cyn} & 0.560 & 4 & \textsc{tool} \\
Llama 2 & 0.374 & 0 & \textsc{tool} & 0.560 & 0 & \textsc{tool} & 0.517 & 0 & \textsc{tool} \\
Stable Beluga 2 & 0.454 & 0 & \textsc{tool} & 0.613 & 0 & \textsc{cyn} & \textbf{0.572} & 0 & \textsc{tool} \\
Command & 0.408 & 0 & \textsc{tool} & 0.592 & 2 & \textsc{cyn}$^{\dagger}$ & 0.538 & 0 & \textsc{tool} \\
GPT 3.5 Turbo 0613 & 0.442 & 0 & \textsc{elt} & \textbf{0.625} & 4 & \textsc{tool} & \textbf{0.572} & 1 & \textsc{tool} \\

\midrule
F: GPT-3.5 Turbo & \multicolumn{1}{c|}{--} & \multicolumn{1}{c|}{--} & \multicolumn{1}{c|}{--} & 0.614 & 0 & \multicolumn{1}{c|}{--} & 0.517 & 0 & \multicolumn{1}{c}{--} \\

F: GPT-3.5 Turbo & \multicolumn{1}{c|}{--} & \multicolumn{1}{c|}{--} & \multicolumn{1}{c|}{--} & 0.597 & 1 & \multicolumn{1}{c|}{--} & 0.531 & 1 & \multicolumn{1}{c}{--} \\

F: GPT-3.5 Turbo & \multicolumn{1}{c|}{--} & \multicolumn{1}{c|}{--} & \multicolumn{1}{c|}{--} & 0.635 & 3 & \multicolumn{1}{c|}{--} & 0.532 & 3 & \multicolumn{1}{c}{--} \\

F: GPT-3.5 Turbo & \multicolumn{1}{c|}{--} & \multicolumn{1}{c|}{--} & \multicolumn{1}{c|}{--} & 0.625 & 5 & \multicolumn{1}{c|}{--} & 0.528 & 5 & \multicolumn{1}{c}{--} \\

L: GPT-3.5 text-davinci-003 & \multicolumn{1}{c|}{--} & \multicolumn{1}{c|}{--} & \multicolumn{1}{c|}{--} & 0.670 & 16 & \multicolumn{1}{c|}{--} & \bf 0.570 & 16 & \multicolumn{1}{c}{--} \\

L: GPT-3.5 text-davinci-003 & \multicolumn{1}{c|}{--} & \multicolumn{1}{c|}{--} & \multicolumn{1}{c|}{--} & \bf 0.693 & 64 & \multicolumn{1}{c|}{--} & \multicolumn{1}{c|}{--} & \multicolumn{1}{c|}{--} & \multicolumn{1}{c}{--} \\

C: GPT-3.5 text-davinci-003 & \multicolumn{1}{c|}{--} & \multicolumn{1}{c|}{--} & \multicolumn{1}{c|}{--} & 0.634 & 2 & \multicolumn{1}{c|}{--} & \multicolumn{1}{c|}{--} & \multicolumn{1}{c|}{--} & \multicolumn{1}{c}{--} \\

C: GPT-4 0314 & \multicolumn{1}{c|}{--} & \multicolumn{1}{c|}{--} & \multicolumn{1}{c|}{--} & 0.650 & 2 & \multicolumn{1}{c|}{--} & \multicolumn{1}{c|}{--} & \multicolumn{1}{c|}{--} & \multicolumn{1}{c}{--}  \\

\midrule

\citet{stahlberg-kumar-2021-synthetic} &\multicolumn{1}{c|}{--}& \multicolumn{1}{c|}{--}& \multicolumn{1}{c|}{--} & \bf 0.647 & \multicolumn{1}{c|}{--}& \multicolumn{1}{c|}{--} & 0.666 &\multicolumn{1}{c|}{--} & \multicolumn{1}{c}{--} \\
\citet{yuan-bryant-2021-document}  & \bf 0.626 & \multicolumn{1}{c|}{--}& \multicolumn{1}{c|}{--} &\multicolumn{1}{c|}{--}& \multicolumn{1}{c|}{--}& \multicolumn{1}{c|}{--} & 0.629 & \multicolumn{1}{c|}{--} & \multicolumn{1}{c}{--} \\
\citet{zhou-et-al-2023-improving-gec} &\multicolumn{1}{c|}{--}& \multicolumn{1}{c|}{--}& \multicolumn{1}{c|}{--} & \multicolumn{1}{c|}{--}& \multicolumn{1}{c|}{--}& \multicolumn{1}{c|}{--} & \bf 0.696 & \multicolumn{1}{c|}{--} & \multicolumn{1}{c}{--} \\



\bottomrule
    \end{tabular}
\caption{\label{table:final_test_results} Results on the FCE, JFLEG, and CoNLL-14 test sets. For each model and test set, we use the prompt that results in the best performance on the corresponding dev set. \textsc{cyn} refers to the prompt from \citet{coyne2023analyzing} with our few-shot examples listed in \autoref{table:few_shot_examples}, while \textsc{cyn}$^{\dagger}$ indicates the prompt from \citet{coyne2023analyzing} with their few-shot examples. Performance reported for GPT$*$ in previous work is shown in the middle part of the table, with the number of few-shot examples where applicable. F: refers to \citet{fang2023chatgpt}, L: to \citet{loem-etal-2023-exploring}, C: to \citet{coyne2023analyzing}. The final section of the table shows SOTA performance by a single non-ensemble system for each test set in the literature. The best scores in each table section are in bold.
}
\end{table*}

\section{Results}
\label{section:zero_few_shot_results}

\autoref{table:final_zero_few_shot_max_dev} presents the top-1 results for each model on the development sets for the FCE, JFLEG and W\&I+LOCNESS. From the models we test, the results show GPT-4 scores highest on every development dataset, though Stable Beluga 2 and GPT-3.5 Turbo obtain comparable performance to GPT-4 on JFLEG. Amongst the open-source models, Falcon-40B-Instruct and Stable Beluga 2 have relatively high performance across the board, whilst Flan-T5 scores highly on FCE dev specifically. 

Contrary to expectations set by previous work, adding few-shot examples to the three zero-shot prompts does not always lead to an improvement in performance.
Indeed for FCE dev, zero-shot prompts perform best for most models. The picture is mixed for JFLEG dev, whilst the majority of models benefit from few-shot learning for W\&I dev. It remains a matter for future work to investigate whether more dynamic approaches to data sampling (as opposed to a fixed selection of examples) will aid with few-shot GEC prompting.

\begin{table}[t]
\footnotesize
\begin{tabular}{lRRR}
\toprule
Error & Falcon & GPT-3.5 & StableB2 \\
\midrule
M:DET & 0.643 & 0.620 & 0.638 \\
M:OTHER & 0.155 & 0.175 & 0.221 \\
M:PREP & 0.447 & 0.403 & 0.422 \\
M:PUNCT & 0.570 & 0.475 & 0.470 \\
R:DET & 0.375 & 0.353 & 0.362 \\
R:MORPH & 0.444 & 0.395 & 0.399 \\
R:NOUN & 0.291 & 0.261 & 0.284 \\
R:NOUN:NUM & 0.633 & 0.570 & 0.593 \\
R:ORTH & 0.597 & 0.609 & 0.589 \\
R:OTHER & 0.281 & 0.300 & 0.296 \\
R:PREP & 0.490 & 0.488 & 0.466 \\
R:PUNCT & 0.365 & 0.503 & 0.315 \\
R:SPELL & 0.781 & 0.769 & 0.761 \\
R:VERB & 0.219 & 0.253 & 0.258 \\
R:VERB:FORM & 0.552 & 0.486 & 0.454 \\
R:VERB:SVA & 0.641 & 0.571 & 0.611 \\
R:VERB:TENSE & 0.499 & 0.471 & 0.516 \\
U:DET & 0.530 & 0.555 & 0.554 \\
\bottomrule
\end{tabular}
\caption{\label{table:wibea_top_k_error_type_perf} F$_{0.5}$ for the 18 most frequent error types in the W\&I+LOCNESS development set, for the 3 best performing models: Falcon-40B-Instruct, GPT-3.5 Turbo, and Stable Beluga 2.}
\end{table}

We find that the four smallest models are biased towards precision over recall, while the larger models are more balanced (\autoref{table:precision_recall_fce_dev}).
The GPT$*$ models have the best recall, which is a finding that deserves further investigation in future work.

\autoref{table:final_test_results} shows the performance of each model (except GPT-4) on the three test sets in our study: FCE, JFLEG and CoNLL-14. We compare with previous work on GEC with LLMs, and SOTA results from GEC-specific systems in the literature.
For FCE and JFLEG, we use the prompt template that resulted in the best performance on the corresponding development set. For example, for GPT-3.5 Turbo on the FCE test set, we use the \textsc{ELT} zero-shot prompt because it resulted in the best performance on FCE dev. For CoNLL-14, we do the same based on model performance for W\&I+LOCNESS dev.

Our LLM results are well short of SOTA performance, established by task-specific supervised models, for FCE test and CoNLL-14 test -- the corpora annotated in minimal edit fashion -- whereas the performance of GPT 3.5 Turbo is much closer to the SOTA on JFLEG test. These findings reinforce initial experiments by \citet{coyne2023analyzing}, \citet{fang2023chatgpt} and \citet{loem-etal-2023-exploring}. It is apparent that supervised GEC systems, trained on each corpus, are best for minimal edit style corrections, whereas LLMs generate SOTA fluency corrections more similar to the style found in JFLEG.





\subsection{Error type analysis}


We use ERRANT to obtain the grammatical error types found in the W\&I+LOCNESS development set -- the largest development set we evaluate. ERRANT can identify 55 error classes. \autoref{table:wibea_top_k_error_type_perf} presents  F$_{0.5}$ scores for the 18 most frequent error types for the three best performing models: Falcon-40B-Instruct, Stable Beluga 2, and GPT-3.5 Turbo.
Performance for each error type is comparable across the models, though GPT-3.5 Turbo is notably better at replacement punctuation errors.

Generally, the LLMs excel at spelling, missing determiners, replacement subject--verb agreement, replacement noun number, and orthography errors, while struggling on the open-class replacement of nouns and verbs, and the catch-all ``other'' error type. It seems that the LLMs perform better on morphological or character-based corrections which are not too distant from the original form, whereas lexical or phrasal replacement within the minimal edit paradigm are much more challenging.

\section{Discussion}

We set out to investigate the performance levels of LLMs on the task of English GEC. Previous work has shown that GPT$*$ models could perform GEC with mixed success: outdoing existing SOTA models on the JFLEG dataset, which contains fluency corrections, whilst performing poorly on benchmarks annotated with minimal edit corrections -- namely CoNLL-14, the FCE and W\&I+LOCNESS \cite{fang2023chatgpt,loem-etal-2023-exploring,coyne2023analyzing}. We aimed to elicit minimal edit corrections through exploration of different prompting strategies, and evaluated models other than GPT$*$ -- including more open-source than commercial LLMs.

Our findings echo those in previous papers: our chosen LLMs perform well on JFLEG test -- above all Falcon-40B-Instruct, Stable Beluga 2 and GPT-3.5 -- though not outdoing SOTA, possibly because our prompts were designed to discourage fluency style corrections. Based on experiments with JFLEG dev, GPT-4 might perform best on JFLEG test, but full investigation of this question requires additional funding as GPT-4 is currently an order of magnitude more expensive than GPT-3.5 Turbo.

In contrast, the LLMs perform poorly on the FCE and CoNLL-14 test sets, lagging far behind SOTA in both cases. For these datasets, open-source models outperform or compete with the commercial models: the best performing model is FLAN-T5 on the FCE, and Stable Beluga 2 matches GPT 3.5 Turbo in the case of CoNLL-14. Again, performance on the FCE and W\&I+LOCNESS dev sets suggests that GPT-4 could outperform the other LLMs on the test sets.


We narrowed down our initial 11 zero-shot prompts to the 7 which performed best on a sample of sentences from the FCE and W\&I+LOCNESS training sets. We created few-shot prompts from the 3 best performing zero-shot prompts and varied the number of examples from 1 to 4. The results for zero-shot versus few-shot learning do not clearly show a best method for prompting. The open-source models which perform best on the test sets are FLAN-T5, Falcon-40B-Instruct, and Stable Beluga 2: of these, FLAN-T5 and Falcon-40B-Instruct work best with few-shot learning, whereas Stable Beluga 2 is best with a zero-shot prompt. For GPT-3.5 Turbo, zero-shot is best for FCE test, few-shot is best for JFLEG and CoNLL-14 test.

In terms of prompt wording, the \textsc{tool} and \textsc{cyn} prompts are best for the three best open-source models: FLAN-T5, Falcon-40B-Instruct, and Stable Beluga 2. For GPT-3.5 Turbo, the \textsc{elt} prompt is best for the FCE test set and the \textsc{tool} one is best for JFLEG and CoNLL-14 test. Note that two of the three best performing prompts are those in which a role is clearly specified to the LLM -- either as an English language teacher or a grammatical error correction tool (\autoref{table:prompts_brief}).

The fact that the other best performing prompt, the one from \citet{coyne2023analyzing}, replicates the strong results from that paper is evidence for convergence around optimal prompt crafting. Further exploration of the huge prompt search space is possible, but we show that the \textsc{cyn} prompt holds up well against a set of alternatives, and can therefore be considered a strong baseline for future GEC experiments.

Another provision we make for replication in future studies is to supply the list of examples we used in few-shot learning (\autoref{table:few_shot_examples}). This allows others to use them for their own novel prompts, while holding constant the nature of the examples. Furthermore we believe that alternative methods for sourcing few-shot examples could be explored in order to dynamically select the most suitable examples for each input sentence, as future work.

Finally, we note that the comparison between commercial and open-source LLMs is not entirely even, as the former sit behind APIs and a black box processing pipeline. We recognise that GPT-3.5 Turbo shows great promise for English GEC, at least for fluency corrections, but we also find that several open-source models perform relatively well -- in fact better than GPT-3.5 on benchmarks annotated with minimal edits. This is a boon for open science, because models which researchers can obtain and work with directly lead to greater transparency in GEC and beyond.

\section{Conclusion}


We have shown that LLMs do not always outperform existing SOTA models for English GEC: for minimal edit style datasets such as the FCE, CoNLL-14 and W\&I+LOCNESS, their performance is far below that of supervised GEC systems. We attempted to elicit minimal edit corrections from LLMs through prompt crafting, but it may be that LLMs are still biased towards fluency rewrites as has been shown in previous work \cite{coyne2023analyzing,fang2023chatgpt,loem-etal-2023-exploring}. This is consistent with our finding, echoing that of others, that LLMs perform best on JFLEG, which was annotated with a fluency correction style.

We arrive at the following conclusions: (i) Supervised models are still best for English GEC with minimal edit corrections; (ii) Further explorations of prompt crafting, few-shot learning, and dynamic sampling are justified, as is work with open-source models as opposed to commercial ones; (iii) Methods for improving LLM performance on specific error types could be explored.

Other potential areas for future work include document-level GEC and human evaluation of proposed corrections. 
We worked with sentence-level GEC, but this deviates from the greater amount of essay context given to annotators. Document-level GEC has been proposed and recommended in previous work \cite{yuan-bryant-2021-document,coyne2023analyzing}. Exploratory work by \citet{fang2023chatgpt} showed that ChatGPT could not perform document-level GEC well, and speculated that it may not be able to handle long inputs requiring ``high levels of coherence and consistency between sentences''.
We notice that LLMs are better at GEC on beginner and intermediate texts, rather than advanced or native-speaker ones (Appendix \autoref{table:wibea_cefr_perf}): further investigation is needed on this matter.


Initial human evaluation studies suggest a preference for the corrections generated by LLMs over the reference corrections contained in GEC corpora \cite{coyne2023analyzing,fang2023chatgpt}. 
It may be that human raters prefer to read the more fluent LLM-derived corrections but minimal edit corrections are actually more helpful for language learning since they are more faithful to the original intended meaning of the writer. Investigating learning benefits from receiving minimal edit grammatical feedback as opposed to fluency rewrites is a matter for future work which will involve longitudinal data collection, a focus on different feedback styles, and tracking how learners respond.




\section*{Ethics statement}

Multiple ethical concerns have been raised with regards to LLMs and their potential for harmful impacts. Without proper safety mechanisms in place LLMs are capable of generating toxic language \cite{wen-etal-2023-unveiling}, displaying bias regarding gender, sexuality and ethnicity \cite{thakur-etal-2023-language,felkner-etal-2023-winoqueer,narayanan-venkit-etal-2023-nationality,wan-etal-2023-kelly}, and failing to represent a diverse range of linguistic varieties \cite{lahoti-etal-2023-improving}. In the context of GEC the possible implications are for over-correction of `errors' which are in fact simply dialect forms, and for proposing biased or even toxic corrections. Evaluating LLMs on established benchmarks as we have done here is one thing, but if end-users are brought into contact with such systems then we would have ethical concerns about the potential for harm through unfiltered responses. Therefore, further research is needed into the safety of such models if used `in the wild' -- for instance through careful user studies and reporting of problems. In this regard, existing GEC systems which do not use generative AI models are safer (and as reported in this paper, they perform minimal edit style correction better).

Another known issue with LLMs is their tendency to at times hallucinate and generate falsehoods \cite{bouyamourn-2023-llms,rawte-etal-2023-troubling,azaria-mitchell-2023-internal}. In a user-facing GEC system this means that responses could include misleading information. As such, further research is needed into ways to mitigate hallucination in LLMs \cite{ji-etal-2023-towards,semnani-etal-2023-wikichat}.

\section*{Limitations}

This work surveyed a wide selection of models and prompts, and compared their performance across four different datasets. As we have noted in Section \ref{section:prompt_templates} (Evaluation), there is an open search space of prompts, and therefore it is not possible to conclude that we have certainly found the best combination of model and prompt and dataset. This is one limitation of investigating prompt-performance -- there might always be a different prompt that results in better performance for a certain model or dataset.

\section*{Acknowledgements}

This paper reports on research supported by Cambridge University Press \& Assessment.
Some of this work was performed using resources provided by the Cambridge Service for Data Driven Discovery (CSD3) operated by the University of Cambridge Research Computing Service, provided by Dell EMC and Intel using Tier-2 funding from the Engineering and Physical Sciences Research Council (capital grant EP/T022159/1), and DiRAC funding from the Science and Technology Facilities Council. We acknowledge the Cohere For AI Research Grant that supported our evaluation of the Cohere Command model.
We thank Yulong Lin for his exploratory work on GEC with LLMs.

\bibliography{anthology,custom}

\newpage

\appendix



\section{Model sizes}

\begin{table}[h]
\centering
\small
\begin{tabular}{lrrrr}
\toprule
Model & Size \\
\midrule
BLOOMZ & 7B \\
Flan-T5 & 11B \\
InstructPalmyra & 20B \\
OPT-IML & 30B \\
Falcon-40b-Instruct & 40B \\
StableBeluga2 & 70B \\
Llama-2 chat & 70B \\
\midrule
Cohere Command & -- \\
OpenAI GPT-3.5 Turbo 0613 & -- \\
OpenAI GPT-4 0613 & -- \\
\bottomrule
\end{tabular}
\caption{\label{table:models} List of models and their approximate sizes.}
\end{table}

\section{Prompt Templates}
\label{app:prompt_templates}

\autoref{app_table:zero_shot_prompts} includes the complete list of zero-shot prompt templates used to perform GEC with LLMs. While the majority of the models use these templates, four models recommend a predefined prompt format -- we describe model-specific prompts below. \autoref{table:few_shot_examples} contains the list of examples used in the few-shot prompts.


\begin{table*}[t]
    \footnotesize
    \centering
    
    \begin{tabular}{llp{0.78\linewidth}}
        \toprule
        Index & Shorthand & Prompt \\
        \midrule
        1 & \textsc{.} & Correct the errors. Do not paraphrase. \\
        2 & \textsc{.} & Grammar. \\
        3 & \textsc{min} & Make minimal changes to the following text such that it is grammatically correct. \\
        4 & \textsc{.} & You are an English language teacher. A student has sent you the following essay. \textbackslash n\{text\}\textbackslash nCorrect the errors in the essay that will best help the student to learn from their mistakes. \\
        5 & \textsc{elt} & You are an English language teacher. A student has sent you the following text. \textbackslash n\{text\}\textbackslash nProvide a grammatical correction for the text, making only necessary changes. Do not provide any additional comments or explanations. If the input text is already correct, return it unchanged. \\
        6 & \textsc{tool} & You are a grammatical error correction tool. Your task is to correct the grammaticality and spelling in the input sentence. Make the smallest possible change in order to make the sentence grammatically correct. Change as few words as possible. Do not rephrase parts of the sentence that are already grammatical. Do not change the meaning of the sentence by adding or removing information. If the sentence is already grammatically correct, you should output the original sentence without changing anything. \textbackslash n\textbackslash nInput sentence: \{text\}\textbackslash nOutput sentence: \\
        7 & \textsc{tool} & You are a grammatical error correction tool. Your task is to correct the grammaticality and spelling in the input sentence. Make the smallest possible change in order to make the sentence grammatically correct. Change as few words as possible. Do not rephrase parts of the sentence that are already grammatical. Do not change the meaning of the sentence by adding or removing information. If the sentence is already grammatically correct, you should output the original sentence without changing anything. \textbackslash n\textbackslash nInput sentence: ``\{text\}''\textbackslash nOutput sentence: `` \\
        8 & \textsc{dn} & Please correct the following text. Do not attempt to rewrite it into perfect English or to interpret the text. Often, things could be expressed better by paraphrase, but the task is to make minimal changes to correct the text.  Do not change anything that is correct.  Please make no changes if there are no errors.  \\
        9 & \textsc{.} & Correct this to standard English: \\
        10 & \textsc{cyn} & Reply with a corrected version of the input sentence with all grammatical and spelling errors fixed. If there are no errors, reply with a copy of the original sentence.\textbackslash n\textbackslash nInput sentence: \{text\}\textbackslash nCorrected sentence:  \\
        11 & \textsc{con} & This sentence is ungrammatical: \{text\}. I would correct the sentence with as few changes as possible like this:  \\
        \bottomrule
    \end{tabular}
    \caption{\label{app_table:zero_shot_prompts} Zero-shot prompts. The prompts without a shorthand were removed after the first evaluation phase on 2,000 trial sentences (\autoref{appendix:filtering_zero_shot_prompts}).}
\end{table*}

\begin{table*}
    \footnotesize
    \centering
    \begin{tabular}{llp{0.8\linewidth}}
    \toprule
    Index &  & Prompt \\ 
    \midrule
    1 & Input & I love this sport. I look forward to the weakened, to go out with my bike and my group of friends. \\
    & Output & I love this sport. I look forward to the weekend to go out with my bike and my group of friends. \\
    2 & Input & Lucy Keyes was the last thriller I've seen. \\
    & Output & Lucy Keyes was the last thriller I saw. \\
    3 & Input & In the biggest cities around the world the traffic nonstop and increase every day. \\
    & Output & In the biggest cities around the world, the traffic is nonstop and increasing every day. \\
    4 & Input & Also, the satisfaction of the customers pushes me to work harder and be better at my job. \\
    & Output & Also, the satisfaction of the customers pushes me to work harder and be better at my job. \\
    \bottomrule
    \end{tabular}
    \caption{\label{table:few_shot_examples} The list of examples used in few-shot prompts. For example, 3-shot prompts include examples, in order, 1, 2, and 3.}
\end{table*}

\subsection{OpenAI GPT-$*$}

We use the Open\-AI ChatCompletion endpoint that formats prompts with separate System and User messages. We adapt our prompts and put the instruction in the System message, and the learner sentence with any ``Input'' tags in the User message. For few-shot prompts, we format each example using separate User and Assistant messages, to mimic a chat-history as context -- see \autoref{table:OpenAI_fewshot_example} for an example.

\subsection{Llama-2-chat}

Llama-2-chat is trained with the following structure for the first turn in chat applications:\footnote{\url{https://huggingface.co/blog/llama2\#how-to-prompt-llama-2}}

\begin{small}
\begin{verbatim}
<s>[INST] <<SYS>>
{system_prompt}
<</SYS>>

{input} [/INST]
\end{verbatim}
\end{small}

We insert the entire GEC instruction into the \texttt{system\_prompt}, and the learner sentence into the \texttt{input}. Where a prompt template uses ``Input:''/``Output:'' tags, we append the output tags after the final \verb|[\INST]|.

For the few-shot prompts, we follow the conversational setup and include examples as:

\begin{small}
\begin{verbatim}
{input 1} [/INST] {hypothesis 1} 
</s><s>[INST] {input 2} [/INST]
\end{verbatim}
\end{small}

\subsection{Stable Beluga 2}

Stable Beluga 2 recommends structuring prompts with \texttt{System}, \texttt{User}, and \texttt{Response} tags:

\begin{small}
\begin{verbatim}
### System:
This is a system prompt, please behave 
and help the user.

### User:
{input}

### Assistant:
{The output of Stable Beluga 2}
\end{verbatim}
\end{small}

\subsection{InstructPalmyra-20B}

InstructPalmyra recommends the following prompt format, including a preamble followed by Instruction, Input, and Response tags:

\begin{small}
\begin{verbatim}
Below is an instruction that describes a 
task, paired with an input that provides
further context. Write a response that 
appropriately completes the request. 
\n\n### Instruction:\n {instruction} 
\n\n### Input:\n {input}\n\n### Response:
\end{verbatim}
\end{small}

\section{Post-processing model output}
\label{appendix:post_processing_details}

For each model, we process the output with the following rules:

\begin{enumerate}
\itemsep0em
    \item Remove ``Output sentence: '', ``Corrected sentence: '', and ``Input sentence: '' from the start of sentences.
    \item Strip double-quotes.
    \item If there is an odd number of quotations, we remove trailing quotations.
    \item For Llama 2, we search for and remove strings from a keyword list (included in \autoref{table:keyword_list_llama}).
    \item For Llama 2, we split model generations based on the keyword list in \autoref{table:keyword_list_llama}.
    \item For Falcon-40B-Instruct, we split model generations based on ``Input sentence:'' -- this mainly impacts the few-shot setting, where the model tends to continue the few-shot pattern and generate a novel learner sentence after the correction.
\end{enumerate}


\begin{table}[t]
    \footnotesize
    \begin{tabular}{lp{0.7\linewidth}}
        \toprule
        Type & Message \\
        \midrule
        System & Reply with a corrected version of the input sentence with all grammatical and spelling errors fixed. If there are no errors, reply with a copy of the original sentence. \\
        User & Input sentence: I think smoke should to be ban in all restarants. \\
        Assistant & Corrected sentence: I think smoking should be banned at all restaurants. \\
        User & Input sentence: We discussed about the issu. \\
        Assistant & Corrected sentence: We discussed the issue. \\
        User & Input sentence: {text} \\
        \bottomrule
    \end{tabular}
    \caption{\label{table:OpenAI_fewshot_example} Example formatting for a few-shot prompt template with Open\-AI's Chat Completion endpoint.}    
\end{table}

\begin{table}[h]
    \small
    \centering
    \begin{tabular}{p{0.8\linewidth}}
    \toprule
        Start of sentence keyword list \\
    \midrule
        "Sure! Here" \\
        "Sure! The sentence" \\
        "Here is a" \\
        "Here's a" \\
        \bottomrule
        \toprule
        Truncation keyword list \\
        \midrule
        "(No changes" \\
        "Explanation:" \\
        "(The corrections" \\
        "(No correction" \\
        "Corrections:" \\
        "Is there anything" \\
        "Here's a list of" \\
        "Here is a list of" \\
        "The original sentence" \\
        "(The original sentence" \\
        "In the original sentence" \\
        "(The sentence" \\
        "(The only error in" \\
        "(Changes made:" \\
        "(The change made" \\
        "(Note: " \\
        "The main issue" \\
        "The only change I made" \\
        "I changed" \\
        "I made.*changes" \\
        \bottomrule
    \end{tabular}
    \caption{\label{table:keyword_list_llama} List of keywords used to clean generations from LLama-2-chat.}
\end{table}

\begin{table}
\centering
\small
\begin{tabular}{lrr}
\toprule
Model & Prompt & F$_{0.5}$ \\
\midrule
BLOOMZ & \textsc{cyn} & 0.259 \\
FLAN-T5 & \textsc{tool} & 0.398 \\
InstructPalmyra & \textsc{elt} & 0.349 \\
OPT-IML & \textsc{tool} & 0.393 \\
Falcon-40B-Instruct & \textsc{tool}$^{\dagger}$ & 0.426 \\
Llama 2 & \textsc{tool} & 0.349 \\
Stable Beluga 2 & \textsc{tool}$^{\dagger}$ & \textbf{0.436} \\
Command & \textsc{cyn} & 0.330 \\
GPT-3.5 Turbo 0613 & \textsc{elt} & 0.434 \\
\bottomrule
\end{tabular}
\caption{\label{table:filter_top1_score} Top-1 performing zero-shot prompt for each model on the sampled development set. Refer to \autoref{app_table:zero_shot_prompts} for the prompts. $^{\dagger}$ indicates the prompt with quotations around the input sentence.}
\end{table}


\section{Filtering zero-shot prompts with a sampled development set}
\label{appendix:filtering_zero_shot_prompts}

We evaluated a long-list of eleven zero-shot prompts with each model on a development set of 2,000 sentences sampled uniformly from the FCE and W\&I+LOCNESS training sets. We report F$_{0.5}$ scores as calculated using the automatic scorer in ERRANT. \autoref{table:filter_top1_score} presents the score for the top-1 performing prompt for each model and prompt-type. 

We find Stable Beluga 2 and GPT-3.5 Turbo perform the best and obtain comparable performance using different prompts: the former using the ``GEC tool'' expert and the latter using the ``English language teacher''. Indeed, we observe that the two expert prompts and the prompt from \citet{coyne2023analyzing} result in the best performance across the models. 

\begin{figure}[t]
     \centering
     \begin{subfigure}[b]{0.45\textwidth}
         \centering
        \includegraphics[width=\textwidth]{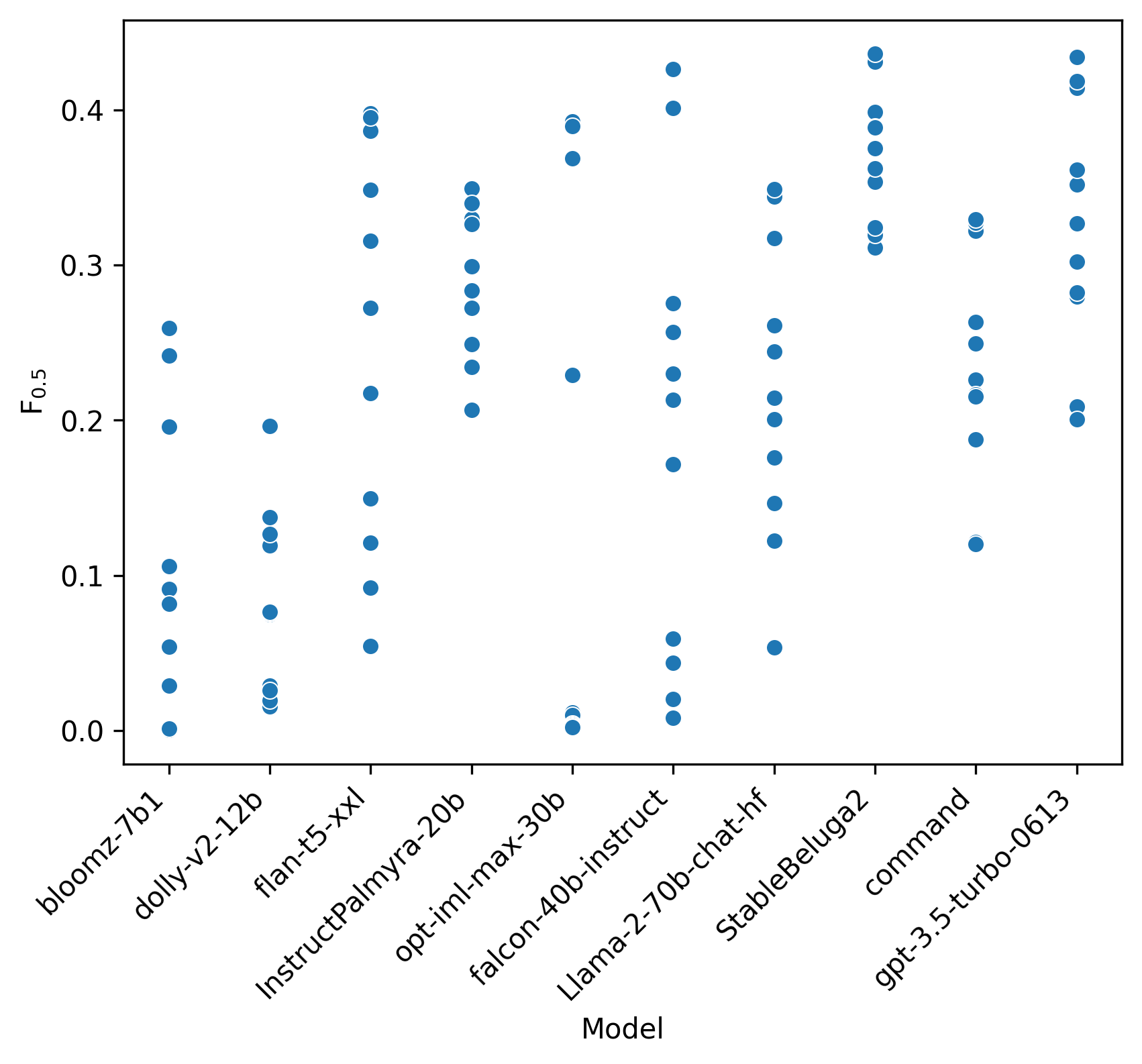}
        \caption{\label{fig:filter_model_f05_scatter} Performance by model.}
     \end{subfigure}
     \hfill
     \begin{subfigure}[b]{0.45\textwidth}
         \centering
        \includegraphics[width=\textwidth]{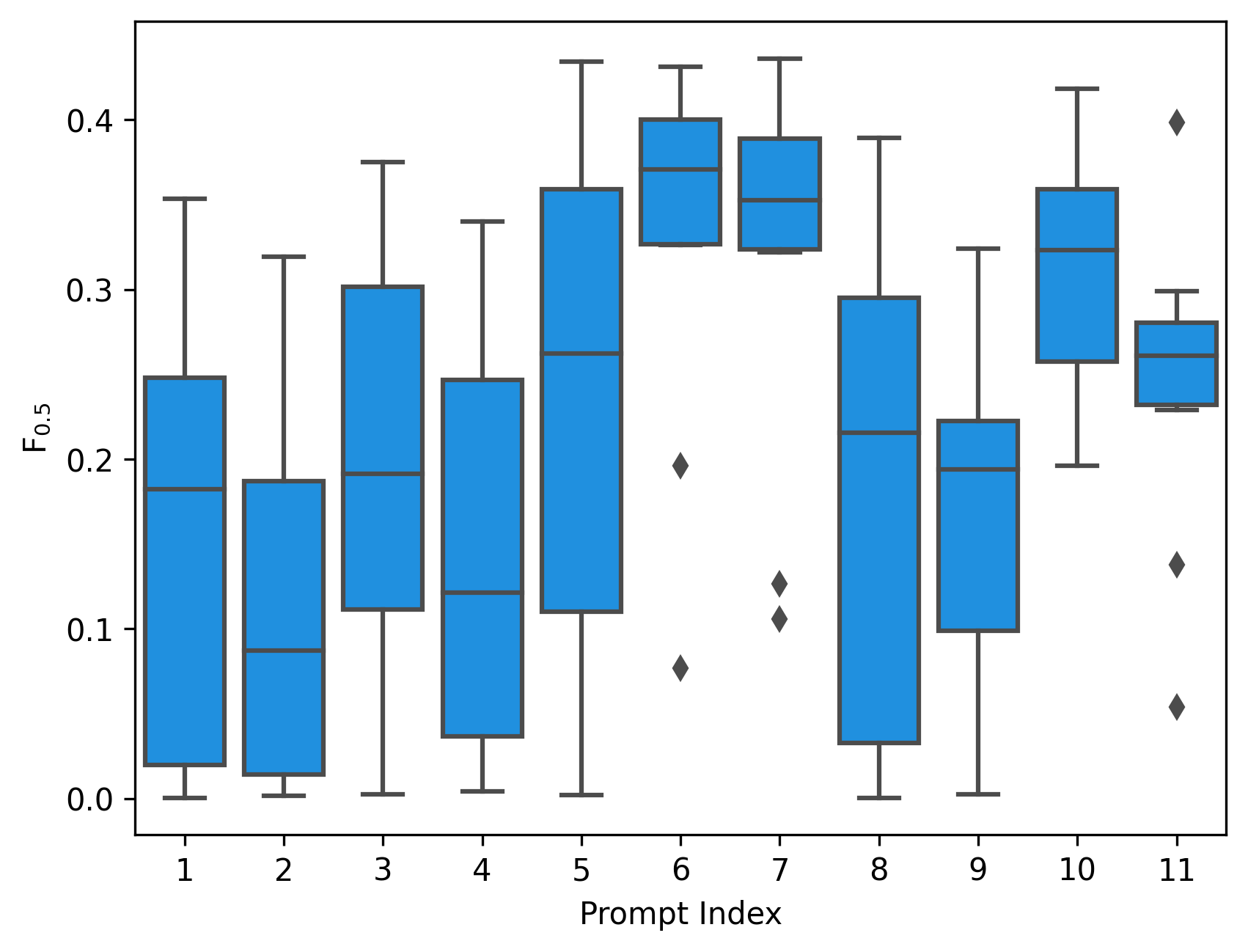}
        \caption{\label{fig:boxplot_f05_zeroshot_prompt_filtering} Performance by prompt.}
     \end{subfigure}
     \caption{\label{fig:perf_zero_shot_sampled_dev} Performance of models using zero-shot prompts on 2,000 sentences sampled uniformly from the FCE and W\&I training sets (1,000 sentences each).}
\end{figure}

\autoref{fig:filter_model_f05_scatter} and \ref{fig:boxplot_f05_zeroshot_prompt_filtering} illustrate F$_{0.5}$ scores for models using the zero-shot prompts, evaluated on the sampled development set. In the former, we can see that Dolly-v2-12B stands out with particularly low performance across all prompts. While in the latter, we can see that prompts 2, 4, and 9 have the lowest maximum scores. Additionally, prompts 4 and 5 are paired: they both use the ``English language teacher'' expert template, but prompt 5 contains more detailed instructions. It is clear from the plot that the more detailed instructions tend to result in higher performance.

From these results, we exclude zero-shot prompts 1, 2, 4, and 9 from the final evaluation due to their relatively low performance with every model. We additionally exclude Dolly-v2-12B due to its low performance across every prompt.

\newpage

\section{Results on the development sets}
\label{appendix:boxplot_dev_sets}

\autoref{table:precision_recall_fce_dev} shows precision, recall and F$_{0.5}$ on the FCE development set. We find that the four smallest models have a bias towards precision over recall, while the larger models are more balanced.

\begin{figure*}
    \centering
    \includegraphics[width=\textwidth]{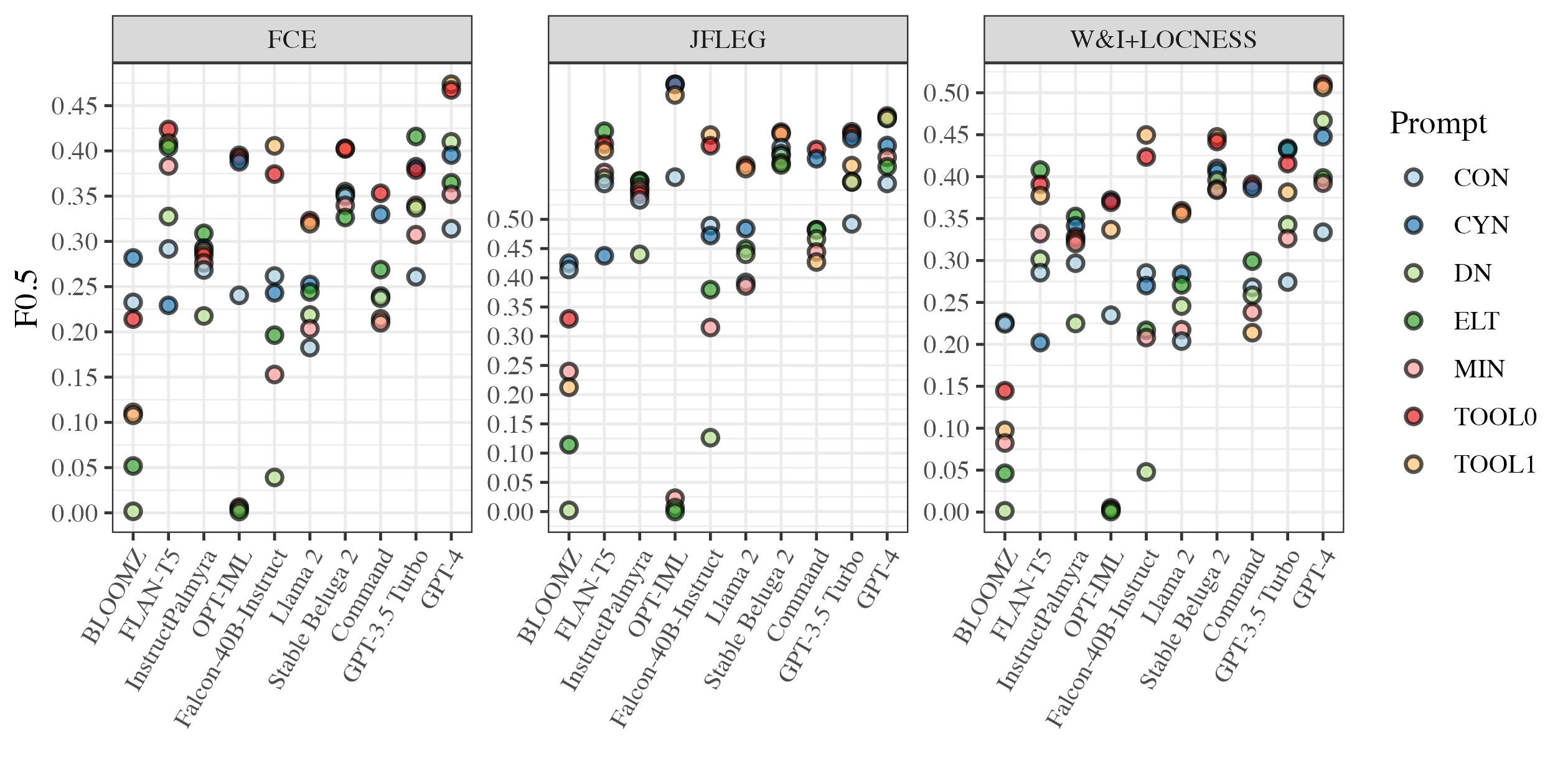}
    \caption{\label{fig:plot_zero_dev} Performance per model and prompt on the FCE development set: F$_{0.5}$ for each model with our seven zero-shot prompts on the FCE, JFLEG and W\&I+LOCNESS development sets. TOOL0 is prompt 6 in \autoref{app_table:zero_shot_prompts} (without quote marks); TOOL1 is prompt 7 (with quote marks). }
\end{figure*}

\begin{figure*}
    \centering
    \includegraphics[width=\textwidth]{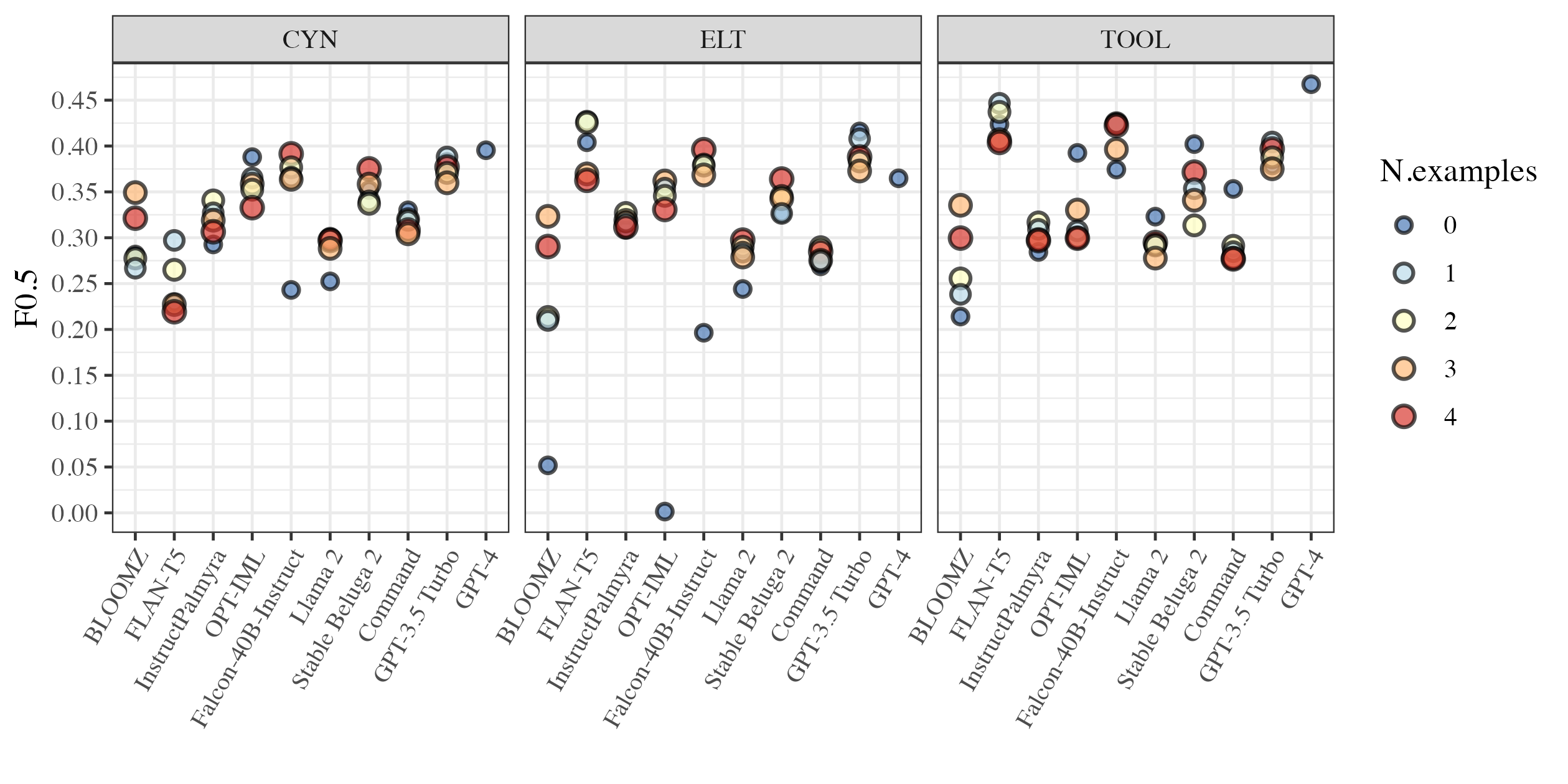}
    \caption{\label{fig:plot_few_dev} Performance per model and prompt on the FCE development set: F$_{0.5}$ for each model with the prompts \textsc{cyn, elt, tool} in zero- and few-shot settings. GPT-4 was only evaluated with zero-shot prompts due to budget constraints.}
\end{figure*}

\autoref{fig:plot_zero_dev} presents the scores for each model on the FCE, JFLEG and W\&I+LOCNESS development sets with our seven zero--shot prompts.
We observe that InstructPalmyra and Stable Beluga 2 have much smaller variance in both zero- and few-shot settings.
On the other hand we observe high variability with different prompts for OPT-IML and Falcon-40B-Instruct. For most models, we observe more consistent performance in the few-shot settings.

\autoref{fig:plot_few_dev} presents the scores for each model on the FCE development set with the prompts \textsc{cyn, elt, tool} in zero- and few-shot settings.
BLOOMZ, OPT-IML, and Falcon-40B-Instruct stand out as particularly sensitive to the choice of prompt -- in particular, OPT-IML scores $\sim$0 F$_{0.5}$ using the \textsc{min}, \textsc{elt}, and \textsc{dn} prompts on each development set.\footnote{OPT-IML generates empty hypotheses for the majority of sentences with prompts \textsc{min}, \textsc{elt}, and \textsc{dn}.}

\begin{table}[t]
\footnotesize
\begin{tabular}{lRRRR}
\toprule
 & \multicolumn{4}{c}{CEFR level} \\

Model & \multicolumn{1}{l}{A} & \multicolumn{1}{l}{B} & \multicolumn{1}{l}{C} & \multicolumn{1}{l}{NS} \\
\midrule
BLOOMZ & 0.349 & 0.328 & 0.328 & 0.396 \\
Flan-T5 & 0.428 & 0.386 & 0.353 & 0.532 \\
InstructPalmyra & 0.408 & 0.375 & 0.280 & 0.388 \\
OPT-IML & 0.421 & 0.359 & 0.325 & 0.486 \\
Falcon-40B-instruct$^{\dagger}$ & 0.487 & 0.465 & 0.373 & 0.434 \\
Llama-2 & 0.412 & 0.380 & 0.273 & 0.315 \\
StableBeluga2$^{\dagger}$ & 0.490 & 0.462 & 0.344 & 0.434 \\
Command & 0.440 & 0.400 & 0.284 & 0.376 \\
GPT-3.5-turbo$^{\dagger}$ & 0.488 & 0.457 & 0.344 & 0.401 \\
GPT-4 & 0.547 & 0.516 & 0.427 & 0.495 \\
\bottomrule
\end{tabular}
\caption{\label{table:wibea_cefr_perf} F$_{0.5}$ for for each proficiency level in the W\&I+LOCNESS development set. $^{\dagger}$ indicates the top 3 performing models for the dataset: Falcon-40B-Instruct, GPT-3.5 Turbo, and Stable Beluga 2. A = beginner learner, B = intermediate, C = advanced, NS = native speaker of English.}
\end{table}

\subsection{Proficiency level analysis}
\label{section:prof_level_results}

We report performance on the W\&I+LOCNESS development set grouped by CEFR level in \autoref{table:wibea_cefr_perf}. The majority of models perform relatively well on A-level learner text (beginners), followed by intermediate B-level text, English text written by native speakers, and finally advanced learner C-level text.

Interestingly, BLOOMZ, FLAN-T5, and OPT-IML perform best on native speaker text. A closer inspection of the precision and recall results show all of these models have a bias towards high precision and low recall.

\end{document}